%% file: main.tex
\documentclass{cameravlm}

\usepackage{amssymb}
\usepackage{tabularx}
\usepackage{xspace}

\title{PhysInsert: Anticipatory Planning and Spatial Preference Optimization for Physics-Verifiable Video Object Insertion}

\author[1,*]{Bohai Gu}
\author[3]{Taiyi Wu}
\author[1]{Dazhao Du}
\author[1]{Jian Liu}
\author[4]{Shuai Yang}
\author[3]{Xiaotong Zhao}
\author[3]{Alan Zhao}
\author[2,*]{Yueyang Yuan}
\author[1]{Xiaoyi Pang}
\author[1]{Jie Zhang}
\author[1,\dagger]{Song Guo}

\affiliation[1]{The Hong Kong University of Science and Technology}
\affiliation[2]{Wuhan University}
\affiliation[3]{AI Technology Center, Tencent Video, Tencent}
\affiliation[4]{Peking University}
\contribution[*]{Equal contribution}
\contribution[\dagger]{Corresponding author}

\abstract{
Video object insertion can produce visually convincing yet physically implausible edits, such as unsupported placement or motion inconsistent with scene forces. We present PhysInsert, an environment-aware framework that couples multimodal large language model (MLLM) reasoning with video diffusion. Its \emph{Anticipatory Spatial Reasoning} anticipates how an insertion should unfold and grounds the resulting plan into frame-wise regions, providing complementary semantic and spatial conditions for generation. Crucially, many physical violations are verifiable from generated videos. PhysInsert turns localized, evidence-based judgments of generated candidates into preference supervision, and uses them to post-train the generator with \emph{Spatial Direct Preference Optimization}. By combining global and insertion-localized objectives, Spatial DPO aligns outcome feedback with the regions where physical violations occur while preserving overall video coherence. PhysInsert also provides flexible and standard modes to balance physically motivated scene adaptation against background preservation. We further introduce complementary benchmarks for insertion with provided regions and for autonomous placement and interaction planning without spatial guidance. Experiments on these benchmarks and the existing UNIC benchmark show improved physical plausibility over the evaluated baselines while maintaining competitive object identity and video quality.
}

\begin{document}
\raggedbottom
\maketitle
\thispagestyle{firstheader}

\input{sec/1_intro}
\input{sec/2_related}
\input{sec/3_method}
\input{sec/4_experiment}
\input{sec/5_conclusion}
\label{main-text-end}
\clearpage
\bibliography{references}
\bibliographystyle{arxiv_refs}
\end{document}

%% file: sec/1_intro.tex
\section{Introduction}
\label{sec:intro}

\begin{figure}[t]
\centering
\includegraphics[width=\linewidth]{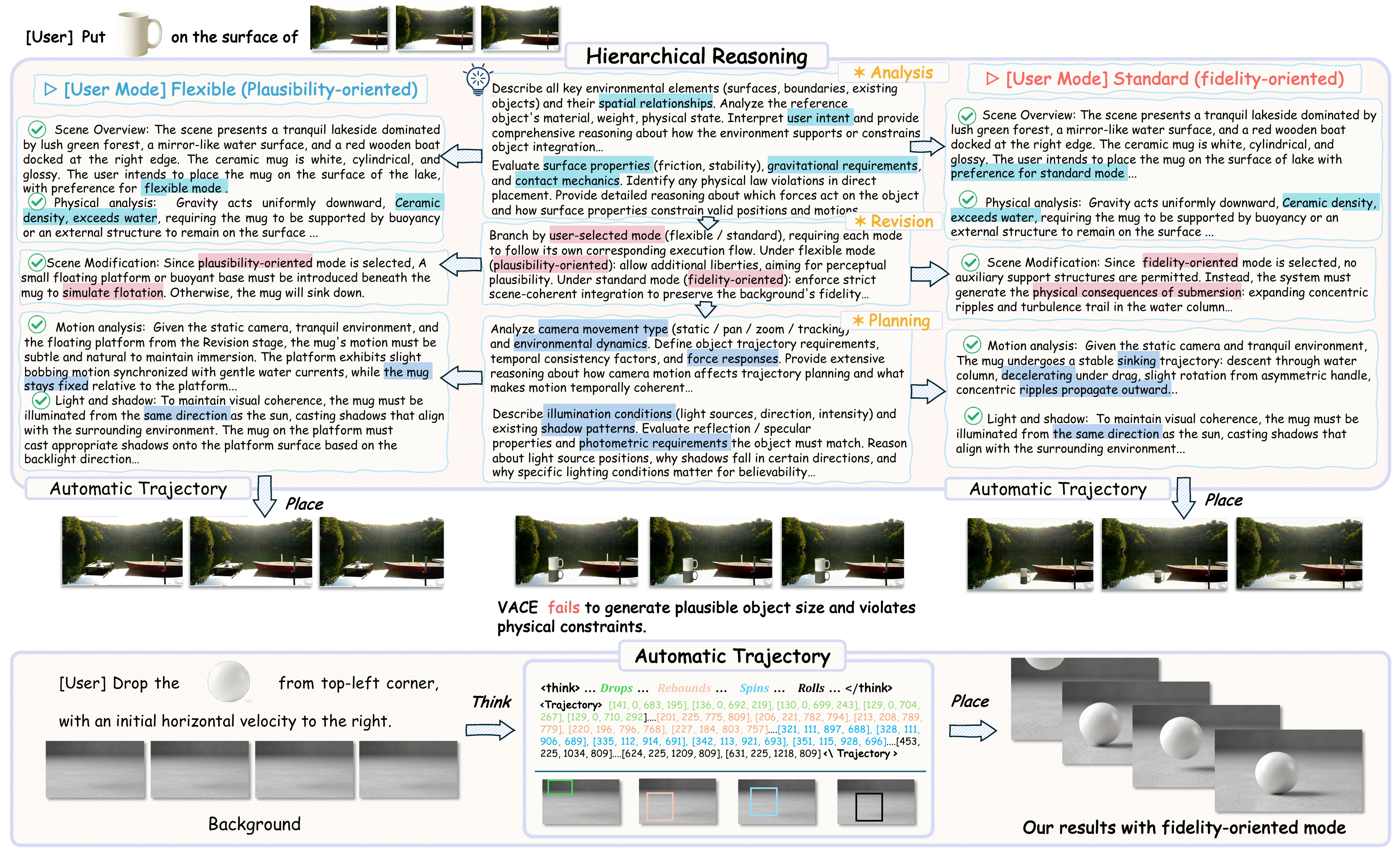}
\caption{PhysInsert performs environment-aware video object insertion through anticipatory spatial reasoning. Its flexible and standard modes provide different trade-offs between physically motivated scene adaptation and background preservation.}
\label{fig:intro}
\end{figure}

Video object insertion places a reference object into a background video according to user instructions~\citep{bai2024anything,saini2024invi,zhao2025dreaminsert}. Recent diffusion-transformer methods~\citep{unic,vace,getinvideo} can produce high-fidelity edits, yet visual realism does not necessarily imply physical plausibility. An inserted mug may float unsupported on water, a ball may pass through the ground, or an object may remain static despite forces in the scene. Existing mask-based systems can reduce placement ambiguity, but they require users to specify frame-wise regions or trajectories. For dynamic interactions, this effectively asks the user to design a physically valid motion path by hand. Inspired by the predictive role of world models, a practical insertion system should instead reason beyond static placement: it should determine \emph{where} an object can plausibly appear and anticipate \emph{how} the insertion will unfold within the environment before rendering begins.

Multimodal large language models (MLLMs) provide a promising source of physical commonsense and scene understanding~\citep{bai2025qwen3vltechnicalreport}, but translating their reasoning into physically plausible video insertion presents three challenges. First, a \emph{representation bottleneck} arises when detailed reasoning is compressed into a short text prompt, discarding object properties, scene constraints, and interaction cues required by the generator. Second, a \emph{spatial grounding gap} remains between a high-level interaction plan and the frame-wise regions needed to execute placement, contact, and motion. Third, reasoning-conditioned generation is inherently open-loop: the plan is produced before rendering and never examines the outcome it induces. This is particularly limiting for physical plausibility because violations such as unsupported placement, incorrect contact, and motion inconsistent with scene forces become verifiable only after the video is generated. Moreover, these violations are typically concentrated around the inserted object, while the surrounding video should remain unchanged. Addressing this outcome-feedback gap therefore requires a post-training mechanism that converts localized assessments of generated results into targeted generator updates without degrading overall video coherence.

We introduce \textbf{PhysInsert}, an end-to-end framework that addresses these challenges through anticipatory reasoning, spatial grounding, and outcome-grounded preference optimization. Its first stage performs \emph{Anticipatory Spatial Reasoning} (ASR), in which an MLLM jointly reasons about insertion geometry and the future physical consequences of placement. ASR analyzes the background, reference object, and user instruction; anticipates how the object should respond to gravity, contact, and scene forces; and predicts frame-wise regions for the object and, when necessary, supporting scene content. This joint treatment of placement and interaction makes the reasoning anticipatory rather than a static description of the edit. PhysInsert then makes the reasoning executable through two complementary pathways: semantic conditioning projects the continuous hidden-state representation of Revision and Planning into the diffusion conditioning space, while spatial conditioning converts predicted trajectories into region masks. The former preserves dense interaction context and communicates \emph{what} should occur and \emph{how} it should unfold; the latter specifies \emph{where} it should be rendered.

The second stage addresses the outcome-feedback gap through physics-verifiable preference post-training. Physical plausibility is particularly amenable to such feedback because many violations can be inspected directly in the generated video. For each input, PhysInsert samples multiple candidates and uses a rubric-based MLLM judge to assess geometric compatibility, appearance integration, and interaction validity with frame-level evidence. The assessment is localized with the predicted insertion regions, and rankings are retained only when repeated evaluations under independently permuted candidate orders agree. The resulting preferred and rejected pairs supervise \emph{Spatial Direct Preference Optimization} (Spatial DPO). Its insertion-localized objective targets the contacts, boundaries, and motions where physical errors occur, while a global objective preserves temporal coherence and the surrounding video. In this sense, PhysInsert \emph{verifies where it optimizes}: the same spatial reasoning that guides generation also anchors outcome assessment and preference learning. This feedback loop is used during training; inference requires only a single pass through the learned planner and generator.

PhysInsert additionally provides two user-selectable modes for different editing requirements. The \emph{flexible mode} permits physically motivated scene adaptation, such as introducing a supporting surface when the requested placement would otherwise be impossible. The \emph{standard mode} prioritizes preservation of the original background and confines adaptation to the inserted object. We further introduce two complementary benchmarks: \textbf{HumanSync}, containing 100 cases with reference insertion regions for controlled execution, and \textbf{FlexInsert}, containing 100 region-free cases for autonomous placement and environment-aware interaction. Evaluation on these benchmarks and 20 cases from the existing UNIC benchmark shows that PhysInsert improves physical plausibility while maintaining competitive identity preservation and video quality.

Our contributions are summarized as follows:
\begin{itemize}
    \item We propose PhysInsert, an environment-aware video object insertion framework built around \emph{Anticipatory Spatial Reasoning}, which jointly predicts object-environment evolution and  physically plausible placement  and translates them into semantic and spatial diffusion conditions.
    \item We introduce physics-verifiable preference post-training with localized, consensus-based candidate assessment and \emph{Spatial DPO}, which couples global alignment with insertion-region optimization under a verify-where-you-optimize principle.
    \item We introduce two complementary benchmarks, \textbf{HumanSync} and \textbf{FlexInsert}, comprising 200 cases that evaluate insertion with provided regions and autonomous placement and interaction planning without spatial guidance. Evaluation on these benchmarks and the existing UNIC benchmark demonstrates improved physical plausibility while maintaining competitive identity preservation and video quality.
\end{itemize}

%% file: sec/2_related.tex
\section{Related Work}
\label{sec:related}

\paragraph{Video editing and object insertion.}
Diffusion models have enabled temporally coherent video editing through training-free propagation, one-shot adaptation, and structured space-time control~\citep{ceylan2023pix2video,geyertokenflow,wu2023tune,liew2023magicedit,mou2024revideo}. Recent general-purpose systems further unify editing conditions and tasks. AnyV2V edits an initial frame and propagates the result through an image-to-video model~\citep{ku2024anyv2v}; VACE introduces modular condition units for diverse video creation tasks~\citep{vace}; and UNIC tokenizes heterogeneous visual conditions for in-context editing~\citep{unic}. Video object insertion places a reference subject into a target video while preserving its identity and temporal consistency~\citep{bai2024anything,saini2024invi}. VideoAnydoor improves subject fidelity and motion control through pixel-level warping~\citep{tu2025videoanydoor}, DreamInsert studies training-free insertion~\citep{zhao2025dreaminsert}, and GetInVideo strengthens temporal modeling with a diffusion-transformer architecture~\citep{getinvideo}. These methods primarily address appearance preservation and controllable placement. PhysInsert instead focuses on anticipating and optimizing the physical relationship between the inserted object and its environment.

\paragraph{Multimodal reasoning and video world models.}
Chain-of-thought reasoning enables language models to decompose complex decisions~\citep{cot}, while recent MLLMs extend such reasoning to visual scenes~\citep{bai2025qwen3vltechnicalreport}. Existing video systems often use multimodal models as semantic encoders or high-level planners~\citep{univideo,uniworld}; however, an executable insertion plan must additionally specify frame-wise spatial support and expected interaction dynamics. Interactive video world models provide a related perspective by predicting scene evolution under actions. WorldCraft expands camera-controllable world models to object-level trajectory manipulation and persistent state~\citep{gu2026worldcraft}. WorldCycle derives self-verifiable supervision from reversible action cycles to improve long-horizon consistency~\citep{gu2026worldcycle}. PhysInsert studies a complementary editing setting: ASR anticipates object-environment evolution before generation, while localized assessment verifies the realized insertion afterward.

\paragraph{Preference optimization for diffusion.}
Preference learning aligns generative models using comparisons between preferred and rejected outputs. Direct Preference Optimization provides a simple alternative to reward-model-based language-model alignment~\citep{rafailov2023direct}, and Diffusion-DPO adapts this principle to image diffusion~\citep{wallace2023diffusion}. VideoDPO extends preference optimization to video generation, while DenseDPO introduces finer temporal supervision~\citep{liu2025videodpo,wu2025densedpo}. Existing approaches generally optimize global generation quality or temporal alignment. Spatial DPO is designed for object insertion, where the most diagnostic differences are concentrated around the inserted object and its contacts. It couples localized, evidence-based assessment with a region-weighted preference objective, while retaining a global term for temporal coherence and background preservation.

%% file: sec/3_method.tex
\begin{figure*}[t]
    \centering
    \includegraphics[width=\textwidth]{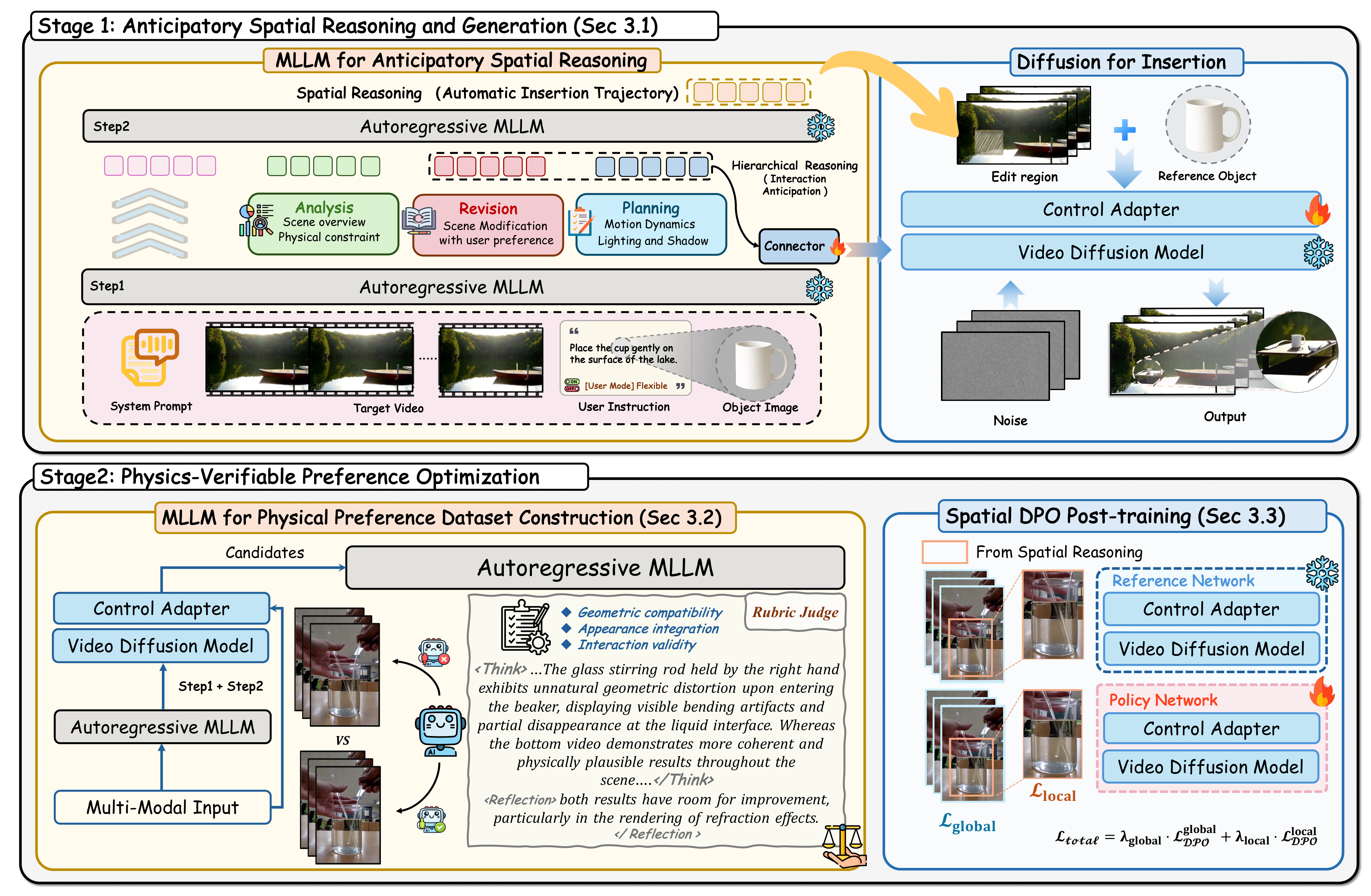}\vspace{-2mm}
    \caption{Overview of PhysInsert. Given a background video, a reference object, and an editing instruction, Stage~1 uses Anticipatory Spatial Reasoning to predict an interaction plan and frame-wise insertion regions. A semantic connector and a spatial control adapter condition the video diffusion model on these complementary signals. Stage~2 constructs a physical preference dataset from generated candidates and applies Spatial DPO, combining global preference alignment with insertion-localized optimization.}
    \vspace{-3mm}
    \label{fig:Overall}
\end{figure*}

\section{PhysInsert Framework}
\label{sec:method}

\subsection{Problem Formulation and Overview}

Let $B=\{b_t\}_{t=1}^{T}$ denote a background video, $O$ a reference image of the object to insert, and $q$ a user instruction. The goal is to generate an edited video $V=\{v_t\}_{t=1}^{T}$ that preserves the identity of $O$ and the unaffected content of $B$, while placing the object at geometrically and physically plausible locations throughout the video. This requires more than appearance transfer: the model must determine a feasible insertion trajectory and render the resulting contacts, motion, lighting, and scene response.

PhysInsert decomposes this problem into planning and outcome alignment (Fig.~\ref{fig:Overall}). In Stage~1, an MLLM performs \emph{Anticipatory Spatial Reasoning} (ASR) to produce (i) a continuous interaction representation $H$ describing how the insertion should unfold and (ii) a sequence of spatial regions $R=\{r_t\}_{t=1}^{T}$ specifying where the object and any permitted scene adaptation should occur. A connector maps $H$ into semantic diffusion conditions, while a control adapter injects the masks derived from $R$. In Stage~2, multiple outputs generated for the same input are compared under explicit, output-verifiable criteria. The resulting preference pairs are used to post-train the generator with Spatial Direct Preference Optimization (Spatial DPO). Thus, ASR specifies a physically informed edit before rendering, whereas Spatial DPO aligns the renderer with the physical quality of its realized outputs.

\subsection{Anticipatory Spatial Reasoning}
\label{sec:3.1}

ASR jointly addresses two coupled questions: \emph{how should the object interact with the scene?} and \emph{where should that interaction occur in each frame?} Treating them separately is problematic: a trajectory is meaningful only under an interaction hypothesis, while a verbal interaction plan is not executable without spatial grounding. We therefore first construct a structured interaction plan and then ground that plan into a frame-wise trajectory.

\subsubsection{Hierarchical Reasoning for Interaction Anticipation}

Given sampled background frames $B$, a reference object $O$, an instruction $q$, and a mode indicator $m\in\{\text{standard},\text{flexible}\}$, the MLLM follows the three-stage reasoning chain shown in Fig.~\ref{fig:Overall}. \textbf{(1) Analysis.} The MLLM first builds a scene overview by identifying relevant surfaces, boundaries, agents, and spatial relations, while examining object properties such as scale, material, and physical state. It then derives the physical constraints implied by support, gravity, contact, and surrounding scene structure, determining what the environment permits before proposing an insertion strategy. \textbf{(2) Revision.} Guided by the user-selected mode, the MLLM reconciles the request with the constraints identified during Analysis. In \emph{flexible mode}, it may revise the scene locally when direct placement is physically infeasible, for example by introducing a supporting surface. In \emph{standard mode}, it preserves the background and instead revises the expected object response, such as submersion, impact, or force-induced motion. This stage determines whether the edit can be executed directly or requires a mode-consistent adaptation. \textbf{(3) Planning.} Conditioned on the analyzed constraints and revised insertion strategy, the MLLM plans the temporal execution of the edit. It accounts for camera motion and environmental dynamics when specifying object motion under gravity, contact, and scene forces, and analyzes illumination, shadow, and reflection for photometric consistency. We retain the hidden states enclosed by dedicated interaction tokens from Revision and Planning as the continuous interaction representation $H$, preserving the contextual reasoning needed by the generator more faithfully than a rewritten text prompt.

\subsubsection{Spatial Reasoning for Automatic Insertion Trajectory}
\label{sec:Automatic Insertion Trajectory}

The second ASR step grounds the interaction plan into frame-wise spatial control. Conditioned on $H$ and the original multimodal inputs, the MLLM first estimates an object scale consistent with scene references and perspective, then selects an initial position with sufficient space for the planned interaction. It subsequently predicts a bounding box
$r_t=[x_t^1,y_t^1,x_t^2,y_t^2]$
for each frame. In flexible mode, the predicted region may also cover localized supporting content introduced during Revision. The temporally linked boxes encode the anticipated trajectory through their displacement and scale variation. We rasterize $R=\{r_t\}_{t=1}^{T}$ into a binary spatiotemporal mask $M\in\{0,1\}^{T\times H_v\times W_v}$, which conditions generation and later localizes preference optimization.

\subsubsection{Reasoning-Conditioned Video Generation}

We instantiate the renderer with a pretrained flow-based video generator equipped with a spatial control branch. Two conditioning pathways translate ASR into executable generation conditions; the concrete backbone and control initialization are specified in Sec.~\ref{sec:implementation}.

\paragraph{Semantic pathway.}
A lightweight connector $g_{\phi}$ projects the interaction representation into the diffusion conditioning space,
\begin{equation}
    C_{\mathrm{sem}} = g_{\phi}(H).
\end{equation}
This condition communicates the intended interaction, motion, and appearance response without reducing the reasoning to a short natural-language prompt.

\paragraph{Spatial pathway.}
The mask $M$, background video $B$, and reference object $O$ are processed by the control adapter to form a spatial condition $C_{\mathrm{spa}}$. It constrains the location and temporal support of the edit, complementing the semantic condition that specifies its content and behavior.

Let $x_1$ be the latent encoding of the target insertion video, $x_0\sim\mathcal{N}(0,I)$, and $x_t=(1-t)x_0+t x_1$ the flow-matching interpolation at time $t\in[0,1]$, with target velocity $u_t=x_1-x_0$. The generator is trained by
\begin{equation}
\mathcal{L}_{\mathrm{FM}}
=\mathbb{E}_{t,x_0,x_1}
\left[
\left\|v_{\theta}(x_t,t;C_{\mathrm{sem}},C_{\mathrm{spa}})-u_t\right\|_2^2
\right].
\label{eq:fm}
\end{equation}
During Stage~1 training, target videos and insertion regions are obtained by reverse-engineering real videos (Sec.~\ref{sec:physinsert}); at inference time, ASR predicts the interaction plan and regions directly from $(B,O,q,m)$.

\subsection{Physics-Verifiable Preference Optimization}
\label{sec:3.2}

ASR provides physically informed conditions, but physical plausibility is ultimately a property of the generated video. Many failures, including unsupported placement, inconsistent scale, incorrect contact, and motion that ignores scene forces, can be identified only after rendering. We therefore turn these output-verifiable cues into preference supervision through candidate assessment followed by Spatial DPO.

\subsubsection{Physical Preference Dataset Construction}

For each conditioning tuple $(B,O,q,m)$, we sample five candidate videos from the Stage~1 generator using different random seeds. An autoregressive MLLM serves as a \emph{rubric judge} and assesses insertion quality along the three criteria shown in Fig.~\ref{fig:Overall}. \emph{Geometric compatibility} measures whether object scale, perspective variation, and spatial relations agree with scene geometry. \emph{Appearance integration} measures whether illumination, shadow, reflection, and their temporal evolution are consistent with the background. \emph{Interaction validity} measures whether support, contact, collision, motion, and causal responses agree with the forces and events visible in the scene. Together, these criteria distinguish an object that is merely recognizable from one that is coherently situated and physically integrated.

For each candidate group, the judge first produces a structured \emph{Think} assessment that compares the candidates under all three rubric criteria and cites frame-level visual evidence. It then produces a \emph{Reflection} that identifies residual weaknesses and checks whether the ranking is supported by the observed geometry, appearance, and interaction. During assessment, the ASR-predicted regions are overlaid to focus attention on the edited content, while the full video remains visible for judging scene-relative geometry and interaction. The resulting rubric-grounded ranking is therefore tied to localized evidence rather than an unexplained holistic score.

Automated rankings can be sensitive to presentation order. To reduce this source of noise, we evaluate every candidate set twice using independently permuted orders and retain it only when the two evaluations produce the same ranking. The highest- and lowest-ranked candidates then form a preferred and rejected pair $(V^{w},V^{l})$. This procedure converts verifiable properties of realized outputs into conservative preference supervision without requiring a differentiable physics metric.

\subsubsection{Spatial Direct Preference Optimization}
\label{sec:3.3}

We initialize a trainable policy $\theta$ and a frozen reference model $\theta_{\mathrm{ref}}$ from the Stage~1 generator. For a video $V$, let $r_{\theta}(V)=v_{\theta}(x_t,t;C)-u_t$ denote the flow-matching residual, where $C=(C_{\mathrm{sem}},C_{\mathrm{spa}})$. The global loss is the mean squared residual over the $N$ elements of the video latent:
\begin{equation}
\ell_{\theta}^{\mathrm{global}}(V)
=
\frac{1}{N}\left\|r_{\theta}(V)\right\|_2^2.
\label{eq:global_fm_loss}
\end{equation}
For each preference pair, the preferred and rejected videos are evaluated using shared timestep and noise sampling; the trainable and reference models also use the same realization. Following the pairwise construction of Diffusion-DPO~\citep{wallace2023diffusion}, we define
\begin{equation}
\Delta^{\mathrm{global}}_{\theta,\mathrm{ref}}
=
\bigl[\ell^{\mathrm{global}}_{\theta}(V^{l})-\ell^{\mathrm{global}}_{\theta}(V^{w})\bigr]
-
\bigl[\ell^{\mathrm{global}}_{\theta_{\mathrm{ref}}}(V^{l})-\ell^{\mathrm{global}}_{\theta_{\mathrm{ref}}}(V^{w})\bigr],
\end{equation}
and optimize
\begin{equation}
\mathcal{L}^{\mathrm{global}}_{\mathrm{DPO}}
=
-\mathbb{E}\left[
\log\sigma\!\left(\beta\Delta^{\mathrm{global}}_{\theta,\mathrm{ref}}\right)
\right].
\label{eq:global_dpo}
\end{equation}
This global term preserves overall temporal coherence and background quality, but it distributes the preference signal uniformly over the video. In object insertion, the most informative differences between preferred and rejected samples are typically concentrated around the inserted object and its interaction boundary.

We therefore define an insertion-localized flow-matching loss using the ASR-predicted spatiotemporal mask. The binary mask is resized to the latent resolution and broadcast over latent channels, yielding $\widetilde M$ with the same shape as the residual. We normalize by its area rather than the full latent size:
\begin{equation}
\ell_{\theta}^{\mathrm{local}}(V,M)
=
\frac{\left\|r_{\theta}(V)\odot\widetilde M\right\|_2^2}
{\|\widetilde M\|_1}.
\label{eq:local_fm_loss}
\end{equation}
Thus, the local term is the mean squared residual over masked latent elements and its scale does not grow with the insertion-region area. Replacing $\ell_{\theta}^{\mathrm{global}}$ by $\ell_{\theta}^{\mathrm{local}}$ in the preference margin gives $\Delta^{\mathrm{local}}_{\theta,\mathrm{ref}}$ and
\begin{equation}
\mathcal{L}^{\mathrm{local}}_{\mathrm{DPO}}
=
-\mathbb{E}\left[
\log\sigma\!\left(\beta\Delta^{\mathrm{local}}_{\theta,\mathrm{ref}}\right)
\right].
\label{eq:local_dpo}
\end{equation}
The final objective balances global preservation and localized correction:
\begin{equation}
\mathcal{L}_{\mathrm{SpatialDPO}}
=
\lambda_{\mathrm{global}}\mathcal{L}^{\mathrm{global}}_{\mathrm{DPO}}
+
\lambda_{\mathrm{local}}\mathcal{L}^{\mathrm{local}}_{\mathrm{DPO}}.
\label{eq:spatial_dpo}
\end{equation}
The region used to focus assessment is thus also the region emphasized during optimization. This shared spatial anchor implements our \emph{verify-where-you-optimize} principle: output feedback is concentrated where the planned insertion is realized, while the global term discourages degradation of the surrounding video.

%% file: sec/4_experiment.tex
\section{Experiments}
\label{sec:experiments}

\subsection{Implementation Details}\label{sec:implementation}

PhysInsert uses Qwen3-VL-8B as the reasoner and Wan2.1-1.3B as the flow-based video generator; its spatial control branch is initialized from VACE-1.3B. We train the two-layer connector and control branch with Eq.~\ref{eq:fm} for 500K iterations on 32 H20 GPUs using AdamW (learning rate $10^{-3}$, batch size 2), while freezing the remaining modules. Spatial DPO uses all 13,000 preference pairs with LoRA rank 128 for 10K iterations and batch size 8, with $\beta=100$ and $\lambda_{\mathrm{global}}=\lambda_{\mathrm{local}}=0.5$. Further optimization, sampling, and data-curation details are provided in the appendix.

\subsection{Data and Benchmarks}
\label{sec:physinsert}
We construct 20,550 Stage~1 reconstruction tuples, of which 18,550 are used for training, and derive 13,000 consensus-filtered preference pairs from the Stage~1 training pool for Spatial DPO. For evaluation, we introduce two complementary benchmarks. \textbf{HumanSync} contains 100 human-object interaction cases with reference insertion regions, enabling controlled evaluation of insertion execution. \textbf{FlexInsert} contains 100 background-video cases without target regions and evaluates autonomous placement, trajectory planning, and environment-aware interaction. We additionally evaluate on 20 cases from the existing UNIC benchmark~\citep{unic} for comparison with prior and commercial systems. All benchmark sources are disjoint from the training data.

\begin{figure*}[t]
    \centering
    \includegraphics[width=0.95\textwidth]{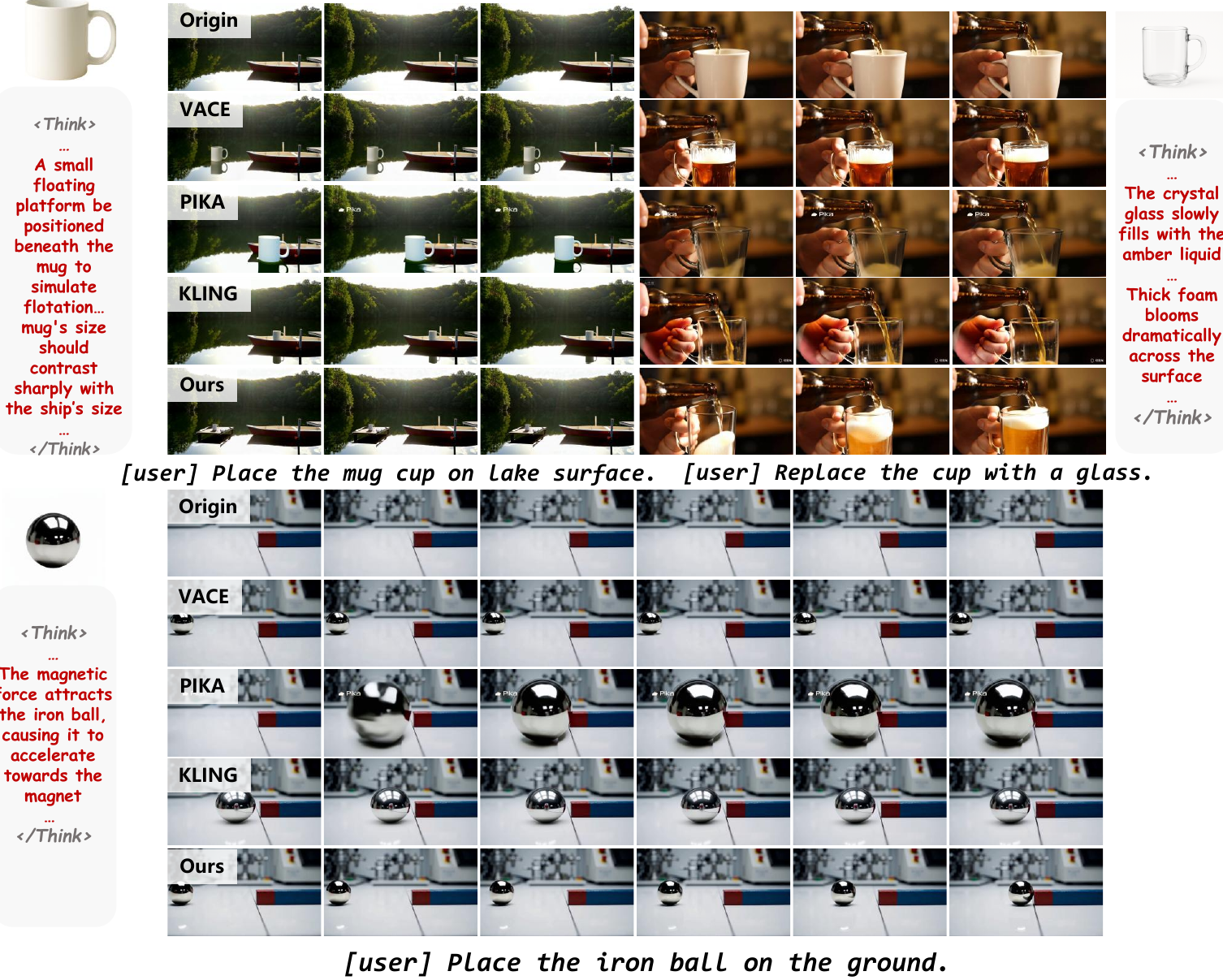}
    \caption{Qualitative comparison of environment-aware object insertion. PhysInsert better preserves plausible object scale, motion, and object-environment interaction across the illustrated scenarios.}
    \vspace{-4mm}
    \label{fig:Qualitative Comparisons.}
\end{figure*}

\begin{table*}[t]
    \centering
    \small
        \caption{
    Quantitative comparisons among three benchmarks. PC: Physical Commonsense, PR: Physical Rule, PP: Physical Plausibility.
    UNIC benchmark includes many virtual animated characters as objects, thus precluding the use of PR.
    }
    \label{table: Quantitative comparison}
    \resizebox{\textwidth}{!}{
        \begin{tabular}{@{}cl*{7}{c}@{}}
        \toprule[1.5pt]
        \multirow{2}{*}{\textbf{Benchmark}} &
        \multirow{2}{*}{\textbf{Method}} &
        \multicolumn{2}{c}{\textbf{Identity}} &
        \multicolumn{2}{c}{\textbf{Video Quality}} &
        \multicolumn{3}{c}{\textbf{Physics}} \\
        \cmidrule(lr){3-4} \cmidrule(lr){5-6} \cmidrule(lr){7-9}
        & & CLIP-I $\uparrow$ & DINO-I $\uparrow$ & Smooth. $\uparrow$ & Aesth. $\uparrow$ & PC $\uparrow$ & PR $\uparrow$ & PP $\uparrow$ \\
        \midrule
\multirow{4}{*}{{\textbf{UNIC}}}
& UNIC~\citep{unic} & 0.5980  & 0.2450 & 0.9610  & 0.5627   & 4.20 & / & 5.33 \\
& Kling (commercial model)~\citep{keling2025elements} & \underline{0.6203}  & 0.2509 & 0.9540  & 0.5641   & 4.41 & / & 5.93 \\
& PIKA (commercial model)~\citep{pika2025}  & \textbf{0.6862}  & \textbf{0.3752} & \textbf{0.9944}  & \textbf{0.6151}   & 4.34 & / & 6.11 \\
& Lucy-edit pro (commercial model)~\citep{Lucy}  & 0.6021  &0.2629 & 0.9865  & 0.5693   & 4.28 & / & 5.79 \\
\cmidrule(lr){2-9}
& \cellcolor{blue!8}\textbf{PhysInsert(standard mode)}    & \cellcolor{blue!8}{0.6043} & \cellcolor{blue!8}\underline{0.2897} & \cellcolor{blue!8}\underline{0.9928} & \cellcolor{blue!8}{0.5684} & \cellcolor{blue!8}\underline{4.53} & \cellcolor{blue!8}{/} & \cellcolor{blue!8}\underline{{6.21}} \\
& \cellcolor{blue!8}\textbf{PhysInsert(flexible mode)}    & \cellcolor{blue!8}{0.6040} & \cellcolor{blue!8}0.2895 & \cellcolor{blue!8}{0.9919} & \cellcolor{blue!8}\underline{0.5787} & \cellcolor{blue!8}\textbf{4.60} & \cellcolor{blue!8}{/} & \cellcolor{blue!8}{\textbf{6.63}} \\
\midrule
\multirow{3}{*}{{\textbf{FlexInsert}}}
& AnyV2V~\citep{ku2024anyv2v} + Anydoor~\citep{chen2024anydoor} & 0.7853  & 0.3805 & 0.9853  & 0.4833   & 3.87 & 0.66 & 3.38 \\
& VACE + Direct Trajectory  & 0.7285  & 0.2541 & \underline{0.9913}  & 0.4920   & 4.03 & 0.67 & 5.21 \\
\cmidrule(lr){2-9}
& \cellcolor{blue!8}\textbf{PhysInsert(standard mode)}   & \cellcolor{blue!8}\textbf{0.7941} & \cellcolor{blue!8}\underline{0.4917} & \cellcolor{blue!8}\textbf{0.9918} & \cellcolor{blue!8}\underline{0.5294} & \cellcolor{blue!8}\underline{4.13}  & \cellcolor{blue!8}\underline{0.78} & \cellcolor{blue!8}\underline{7.28} \\
& \cellcolor{blue!8}\textbf{PhysInsert(flexible mode)}   & \cellcolor{blue!8}\underline{0.7938} & \cellcolor{blue!8}\textbf{0.4925} & \cellcolor{blue!8}{0.9906} & \cellcolor{blue!8}\textbf{0.5305} & \cellcolor{blue!8}\textbf{4.17}  & \cellcolor{blue!8}\textbf{0.86} & \cellcolor{blue!8}\textbf{7.93} \\
\midrule
\multirow{3}{*}{\textbf{HumanSync}}
& VACE~\citep{vace}    & 0.7553  & 0.4210 & 0.9908  & 0.4952   & 4.12 & \underline{0.91} &6.21  \\
\cmidrule(lr){2-9}
& \cellcolor{blue!8}\textbf{PhysInsert(standard mode)}    & \cellcolor{blue!8}\underline{0.7631} & \cellcolor{blue!8}\underline{0.4497} & \cellcolor{blue!8}\textbf{0.9929} & \cellcolor{blue!8}\underline{0.5283}  & \cellcolor{blue!8}\underline{4.33}  & \cellcolor{blue!8}\textbf{0.92} & \cellcolor{blue!8}\underline{6.58} \\
& \cellcolor{blue!8}\textbf{PhysInsert(flexible mode)}   & \cellcolor{blue!8}\textbf{0.7632} & \cellcolor{blue!8}\textbf{0.4500} & \cellcolor{blue!8}\underline{0.9926} & \cellcolor{blue!8}\textbf{0.5295}  & \cellcolor{blue!8}\textbf{4.37}  & \cellcolor{blue!8}\textbf{0.92} & \cellcolor{blue!8}\textbf{6.93} \\
        \bottomrule[1.5pt]
        \end{tabular}
    }
        \vspace{-4mm}
\end{table*}

\begin{table*}[t]
    \centering
    \caption{Ablations on FlexInsert. Left: hierarchical reasoning and conditioning pathways in Stage~1. All variants use the same Spatial DPO post-training. Right: preference optimization in Stage~2. All variants use the same complete Stage~1 model with continuous semantic and spatial conditioning. The data, backbone, and remaining training settings are fixed within each panel.}
    \label{table: Ablation study}
    \begin{minipage}[t]{0.59\textwidth}
        \centering
        \textbf{(a) ASR conditioning}\par\vspace{1mm}
        \resizebox{\linewidth}{!}{
        \begin{tabular}{@{}l*{6}{c}@{}}
        \toprule[1.5pt]
        \multirow{2}{*}{\textbf{Variant}} &
        \multicolumn{2}{c}{\textbf{Identity}} &
        \multicolumn{2}{c}{\textbf{Video Quality}} &
        \multicolumn{2}{c}{\textbf{Physics}} \\
        \cmidrule(lr){2-3} \cmidrule(lr){4-5} \cmidrule(lr){6-7}
        & CLIP-I $\uparrow$ & DINO-I $\uparrow$ & Smooth. $\uparrow$ & Aesth. $\uparrow$ & PC $\uparrow$ & PR $\uparrow$ \\
        \midrule
        Direct trajectory (without ASR) & 0.7305 & 0.3747 & \textbf{0.9923} & 0.5137 & 4.05 & 0.70 \\
        ASR: spatial only & 0.7678 & 0.4489 & 0.9862 & 0.4989 & 3.92 & 0.67 \\
        ASR: text + spatial & 0.7832 & 0.4492 & 0.9892 & 0.5102 & 4.02 & 0.69 \\
        \cellcolor{blue!8}\textbf{Full ASR: continuous + spatial} & \cellcolor{blue!8}\textbf{0.7938} & \cellcolor{blue!8}\textbf{0.4925} & \cellcolor{blue!8}0.9906 & \cellcolor{blue!8}\textbf{0.5305} & \cellcolor{blue!8}\textbf{4.17} & \cellcolor{blue!8}\textbf{0.86} \\
        \bottomrule[1.5pt]
        \end{tabular}}
    \end{minipage}
    \hfill
    \begin{minipage}[t]{0.39\textwidth}
        \centering
        \textbf{(b) Preference optimization}\par\vspace{1mm}
        \resizebox{\linewidth}{!}{
        \begin{tabular}{@{}l*{6}{c}@{}}
        \toprule[1.5pt]
        \multirow{2}{*}{\textbf{Variant}} &
        \multicolumn{2}{c}{\textbf{Identity}} &
        \multicolumn{2}{c}{\textbf{Video Quality}} &
        \multicolumn{2}{c}{\textbf{Physics}} \\
        \cmidrule(lr){2-3} \cmidrule(lr){4-5} \cmidrule(lr){6-7}
        & CLIP-I $\uparrow$ & DINO-I $\uparrow$ & Smooth. $\uparrow$ & Aesth. $\uparrow$ & PC $\uparrow$ & PR $\uparrow$ \\
        \midrule
        Without DPO & 0.7721 & 0.4548 & 0.9891 & 0.4936 & 4.09 & 0.75 \\
        Global DPO & 0.7903 & 0.4693 & 0.9901 & 0.5283 & 4.10 & 0.73 \\
        \rowcolor{blue!8}\textbf{Spatial DPO} & \textbf{0.7938} & \textbf{0.4925} & \textbf{0.9906} & \textbf{0.5305} & \textbf{4.17} & \textbf{0.86} \\
        \bottomrule[1.5pt]
        \end{tabular}}
    \end{minipage}
    \vspace{-3mm}
\end{table*}

\subsection{Comparison with State-of-the-Art Methods}

\subsubsection{Quantitative comparisons.}

We evaluate PhysInsert on two proposed benchmarks and one existing benchmark. \textbf{HumanSync} contains 100 human-object interaction cases with reference insertion regions, enabling controlled evaluation of insertion execution. We compare with VACE~\citep{vace}, providing it with a concise interaction prompt. \textbf{FlexInsert} contains 100 background-video cases without target regions and evaluates autonomous placement, trajectory planning, and environment-aware interaction. On this benchmark, we compare with VACE and AnyV2V~\citep{ku2024anyv2v} combined with Anydoor~\citep{chen2024anydoor}; VACE receives insertion regions predicted directly from the original multimodal inputs without hierarchical ASR. We additionally evaluate on 20 cases from the existing UNIC benchmark~\citep{unic}, comparing with the closed-source UNIC model and the commercial systems Lucy-Edit pro~\cite{Lucy}, Pika~\cite{pika2025}, and Kling~\cite{keling2025elements}.
We evaluate across three dimensions: Identity Preservation (CLIP-I~\citep{radford2021learning} and DINO-I~\citep{caron2021emerging}), Video Quality~\citep{huang2024vbench} (temporal smoothness and aesthetics), and Physical metrics measured using the VideoPhY2 benchmark~\citep{bansal2025videophy}, which quantitatively assesses Physical Commonsense (PC) and Physical Rules (PR) adherence. We further introduce Gemini Pro~\citep{gemini} to evaluate Physical Plausibility (PP) of generated videos.
As shown in Table~\ref{table: Quantitative comparison}, PhysInsert consistently achieves the best performance in physical realism metrics while maintaining competitive performance on video quality and identity metrics. Specifically, the PC~\citep{bansal2025videophy} and PR~\citep{bansal2025videophy} scores show substantial improvements over VACE (6.07\%) and UNIC (9.52\%).  Notably, we conduct a systematically designed human evaluation to assess physical plausibility and video quality on the Flexinsert benchmark with 10 independent annotators who evaluated different tasks in Appendix.

\begin{figure*}[t]
    \centering
    \begin{tabular}{@{}cc@{}}
    \includegraphics[width=0.43\textwidth]{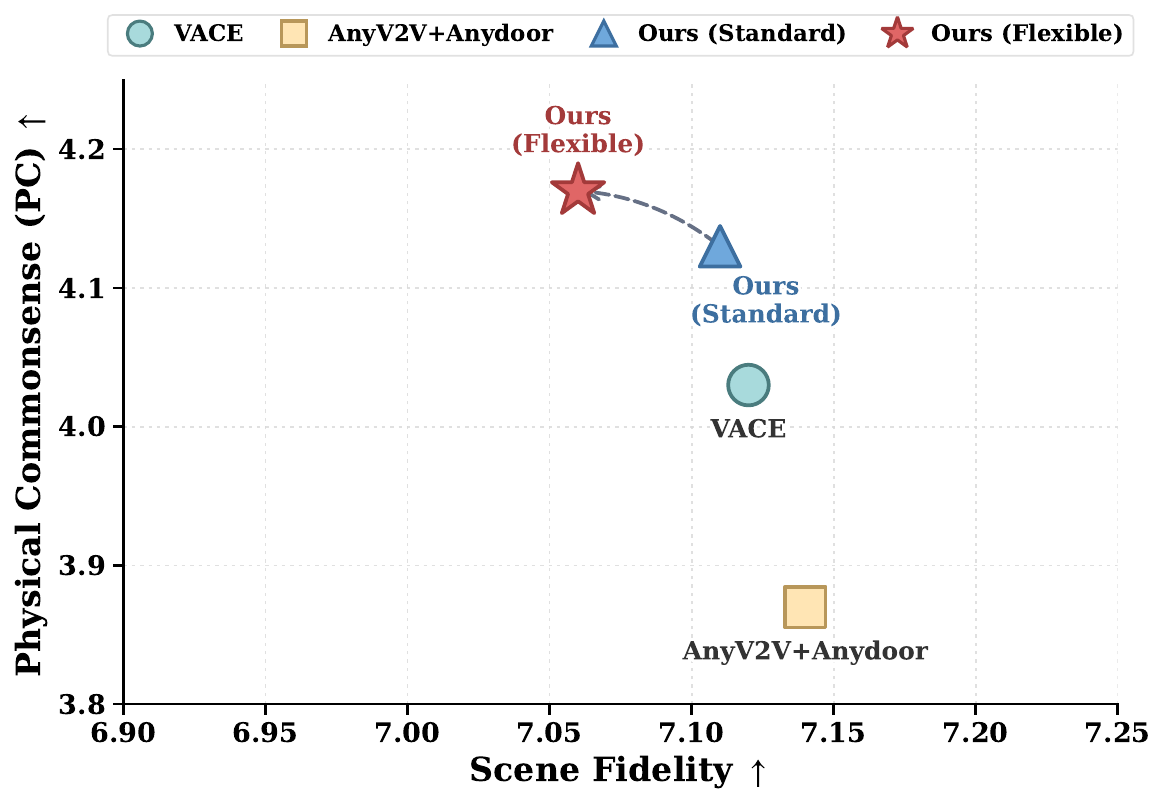} &
    \includegraphics[width=0.50\textwidth]{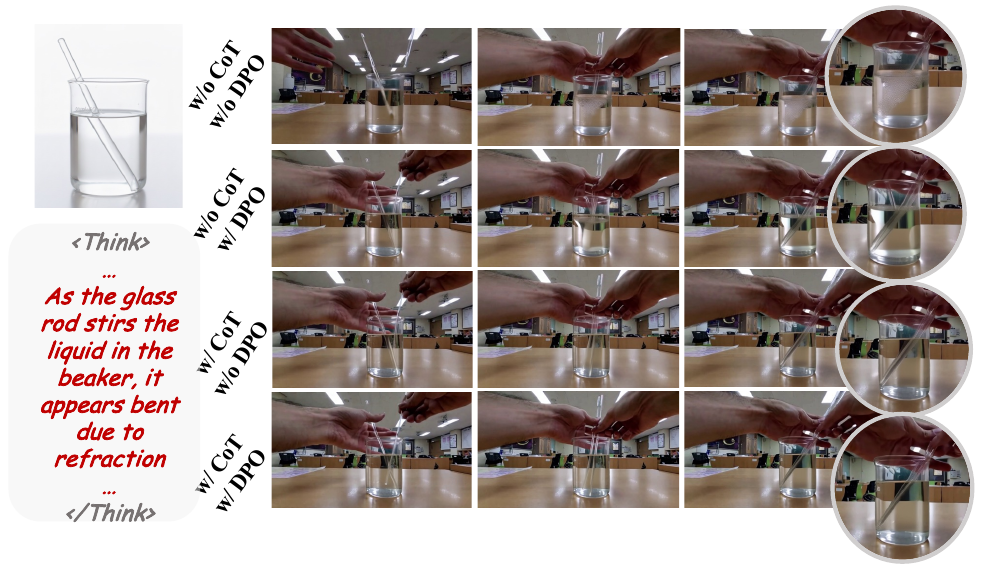}
    \end{tabular}
    \caption{Analysis of PhysInsert. Left: the two modes expose a trade-off between physical plausibility and scene fidelity. Right: anticipatory reasoning and Spatial DPO provide complementary gains.}
    \label{fig:compact_ablations}
    \vspace{-3mm}
\end{figure*}

\subsubsection{Qualitative Comparisons.}

We compare PhysInsert (flexible mode) with VACE~\citep{vace}, Kling~\citep{keling2025elements} and pika~\cite{pika2025}. Since UNIC is not open-source, we utilize their demos for comparisons in Appendix. For a controlled comparison, we provide the mask-based VACE model with a concise interaction prompt and insertion regions predicted directly from the original multimodal inputs without hierarchical ASR.
As demonstrated in Fig.~\ref{fig:Qualitative Comparisons.}, PhysInsert shows superior performance in video object insertion that are not only visually coherent but also physically plausible.
\textbf{Top-Left} (Physics-based interactions): When tasked with placing a mug on a lake's surface, baseline methods fail to adhere to basic physical laws, either placing the mug directly on the water or incorrectly on the boat at an unrealistic scale. In contrast, PhysInsert exhibits strong commonsense reasoning. It correctly infers the need for a support structure, generating a floating platform to realistically simulate flotation while ensuring the mug's size is reasonable relative to the boat.
\textbf{Top-Right} (Fluid Dynamics): In a challenging object-swap task where a cup is replaced with a glass during pouring beer, only our method successfully models the fluid dynamics. It realistically depicts the beer filling the glass and overflowing with authentic foam formation. Other methods fail to capture these dynamic properties, resulting in static or physically inconsistent outcomes.
\textbf{Bottom} (Implicit Force Reasoning): The advantage of our MLLM-guided reasoning is most evident in the bottom scenario. Given the instruction to place an iron ball in an environment with a magnet, PhysInsert is the only method that correctly interprets the unseen magnetic force. It vividly renders the ball accelerating towards the magnet, demonstrating a deep understanding of the scene's underlying physics. Competing methods fail to recognize this crucial context, generating a ball of incorrect size and with implausible motion. These results validate PhysInsert's extraordinary capability in achieving both visual fidelity and physical plausibility in complex, dynamic scenarios.

\subsection{Ablation Study}\label{sec:ablation}

We conduct ablation studies under the flexible setting of PhysInsert. Table~\ref{table: Ablation study} separates the two training stages. In panel (a), all variants undergo the same Spatial DPO post-training, and only the Stage~1 conditioning pathways are changed. In panel (b), all variants start from the same complete Stage~1 model with continuous semantic and spatial conditioning, and only the Stage~2 preference objective is changed.

\noindent{\textbf{\textit{Semantic and spatial conditioning.}}}
Panel (a) progressively exposes the generator to different outputs of ASR. \emph{Direct trajectory (without ASR)} bypasses Analysis, Revision, and Planning, predicting insertion regions directly from the original multimodal inputs. \emph{ASR: spatial only} uses the hierarchical plan to guide trajectory prediction, but conditions the generator only on the resulting region masks. \emph{ASR: text + spatial} additionally conveys the interaction plan through the text encoder. \emph{Full ASR: continuous + spatial} instead projects the continuous Revision and Planning representation through the connector and combines it with ASR-guided spatial control. This progression isolates the benefit of anticipatory trajectory planning and the additional value of preserving its continuous semantic representation.

\noindent{\textbf{\textit{Preference optimization.}}}
Panel (b) fixes the complete Stage~1 pipeline. Without DPO evaluates the reconstruction-trained generator directly. Global DPO applies preference alignment over the complete video, where localized interaction errors can be diluted by the substantially larger background. Spatial DPO supplements this global alignment with an insertion-region objective, producing stronger physical metrics while preserving identity and scene-wide temporal coherence.

\noindent{\textbf{\textit{Plausibility-fidelity trade-off.}}}
To quantify the plausibility-fidelity trade-off between the two modes, we evaluate on the FlexInsert benchmark, using Gemini Pro to assess scene fidelity, which measures the degree of background  preservation relative to the original video. As shown in Fig.~\ref{fig:compact_ablations} (left), the standard mode preserves higher scene fidelity by faithfully maintaining the original background, while  the flexible mode achieves notably stronger physical plausibility by adaptively modifying the environment when physics demands it.  They both offer complementary strengths along different axes, allowing users to freely select the appropriate mode based on their specific editing priorities.

\noindent{\textbf{\textit{Complementarity of reasoning and preference optimization.}}}
Figure~\ref{fig:compact_ablations} (right) provides a qualitative decomposition in a challenging human-interaction scenario. Without anticipatory reasoning and Spatial DPO, the result exhibits temporal and boundary artifacts. Spatial DPO improves rendering naturalness but cannot recover an interaction absent from the plan, whereas reasoning without DPO establishes the intended interaction but retains optical artifacts. Full PhysInsert combines both strengths, producing coherent insertion, fluid behavior, and refraction.

%% file: sec/5_conclusion.tex
\section{Conclusion}
\label{sec:conclusion}
We presented PhysInsert, an environment-aware framework for physically plausible video object insertion. PhysInsert separates anticipation from outcome alignment. Anticipatory Spatial Reasoning analyzes scene constraints, plans object-environment evolution, and translates the plan into complementary continuous semantic and frame-wise spatial conditions. Physics-verifiable preference construction then assesses generated candidates through geometric compatibility, appearance integration, and interaction validity, while Spatial DPO combines insertion-localized correction with global preservation. Experiments on HumanSync, FlexInsert, and UNIC demonstrate improved physical plausibility while maintaining competitive object identity and video quality. The standard and flexible modes further provide practical control over the trade-off between preserving the original scene and adapting it when physical plausibility requires localized changes. More broadly, PhysInsert shows how multimodal reasoning can be grounded before generation and verified through realized outcomes during post-training.

\paragraph{Limitations.}
PhysInsert remains bounded by the perception and physical commonsense of its MLLM reasoner and rubric judge. Unusual materials, subtle forces, severe occlusion, and interactions whose causes or consequences extend beyond the observed clip may lead to incomplete plans or uncertain preference judgments. The current trajectory representation relies on frame-wise regions and therefore does not explicitly model object geometry, depth, or articulated contact. Although Spatial DPO concentrates supervision on insertion regions, model parameters are shared across the video, so localized updates may still influence background appearance or temporal coherence outside the mask. Flexible mode can introduce supporting scene content, but the resulting adaptation is visually plausible rather than guaranteed to be physically correct. Finally, our metrics and human evaluations measure perceived plausibility and consistency; they do not constitute formal physical verification. Future work could incorporate explicit 3D or dynamics representations, uncertainty-aware assessment, longer temporal contexts, and multi-object interaction modeling while preserving the single-pass inference design.